%% file: main.tex
\documentclass[sigconf]{acmart}
\usepackage{algorithm}
\usepackage{algpseudocode}
\usepackage{booktabs}
\usepackage{array}
\usepackage{multirow}   
\usepackage{makecell}
\usepackage{graphicx}
\usepackage{subcaption}
\usepackage{balance}
\usepackage{flushend}
\usepackage{tabularx}
\usepackage{colortbl}   
\usepackage{xcolor}  
\usepackage{caption}
\usepackage{booktabs}

\AtBeginDocument{%
  }

\setcopyright{acmlicensed}
\copyrightyear{2018}
\acmYear{2018}
\acmDOI{XXXXXXX.XXXXXXX}
\setcopyright{rightsretained}
\copyrightyear{2026}
\acmYear{2026}
\acmDOI{10.1145/XXXXXXX.XXXXXXX} 
\acmConference[]{}{}{}
\acmISBN{978-1-4503-XXXX-X/26/03}

\acmISBN{978-1-4503-XXXX-X/2018/06}

\begin{document}

\title{MultiCheck: Strengthening Web Trust with Unified Multimodal Fact Verification}
\settopmatter{printacmref=false}
\renewcommand\footnotetextcopyrightpermission[1]{}

\author{Aditya Kishore}
\email{adityak21@iiserb.ac.in}
\affiliation{%
  \institution{IISER Bhopal}
  \country{}
}

\author{Gaurav Kumar}
\email{gaurav22@iiserb.ac.in}
\affiliation{%
  \institution{IISER Bhopal}
  \country{}
}

\author{Jasabanta Patro}
\email{jpatro@iiserb.ac.in}
\affiliation{%
  \institution{IISER Bhopal}
  \country{}
}

\renewcommand{\shortauthors}{Trovato et al.}

\input{00Abstract}





\maketitle

\input{10introduction}
\input{20RelatedWorks}
\input{30Dataset}
\input{40Methods}

\input{50Experiments}
\input{60Results}
\input{70ErrorAnalysis}

\input{80Conclusion}
\balance
\bibliographystyle{ACM-Reference-Format}
\bibliography{software}
\input{90appendix}

\end{document}

%% file: 00Abstract.tex
\begin{abstract}
Misinformation on the web increasingly appears in multimodal forms, combining text, images, and OCR-rendered content in ways that amplify harm to public trust and vulnerable communities. While prior fact-checking systems often rely on unimodal signals or shallow fusion strategies, modern misinformation campaigns operate across modalities and require models that can reason over subtle cross-modal inconsistencies in a transparent and responsible manner. We introduce \textbf{MultiCheck}, a lightweight and interpretable framework for multimodal fact verification that jointly analyzes textual, visual, and OCR evidence. At its core, MultiCheck employs a relational fusion module based on element-wise difference and product operations, allowing for explicit cross-modal interaction modeling with minimal computational overhead. A contrastive alignment objective further helps the model distinguish between supporting and refuting evidence while maintaining a small memory and energy footprint, making it suitable for low-resource deployment. Evaluated on the Factify-2 (5-class) and Mocheg (3-class) benchmarks, MultiCheck achieves huge performance improvement and remains robust under noisy OCR and missing modality conditions. Its efficiency, transparency, and real-world robustness make it well-suited for journalists, civil society organisations, and web integrity efforts working to build a safer and more trustworthy web.

\end{abstract}

%% file: 10introduction.tex
\vspace{-1mm}
\section{Introduction:}

Misinformation has emerged as a critical challenge in today’s digital ecosystem, with significant impacts across domains such as politics, public health, and finance \cite{caceres2022impact}. While early misinformation was primarily text-based \cite{murphy2023we, kim2021systematic, di2021fake}, recent campaigns increasingly blend text with images, audio, and video, making false claims more persuasive and harder to detect \cite{abdali2024multi, mura2025fake, askari2023deepfakes}. This surge in multimodal misinformation exposes the limitations of traditional fact-checking systems, which focus mainly on textual content \cite{tufchi2023comprehensive, braun2024defame, mura2025fake}. As a result, the research community has shifted toward multimodal fact-checking, where claims are verified using both textual and visual signals \cite{akhtar2023multimodal, braun2024defame}. Benchmarks such as FEVER~\cite{thorne-etal-2018-fever}, Mocheg~\cite{yao2023mocheg}, and Factify 2~\cite{suryavardan2023factify} have accelerated progress in this direction (see Section~\ref{related_work_1}).


\begin{figure}[ht]
\centering
\includegraphics[width=\columnwidth]{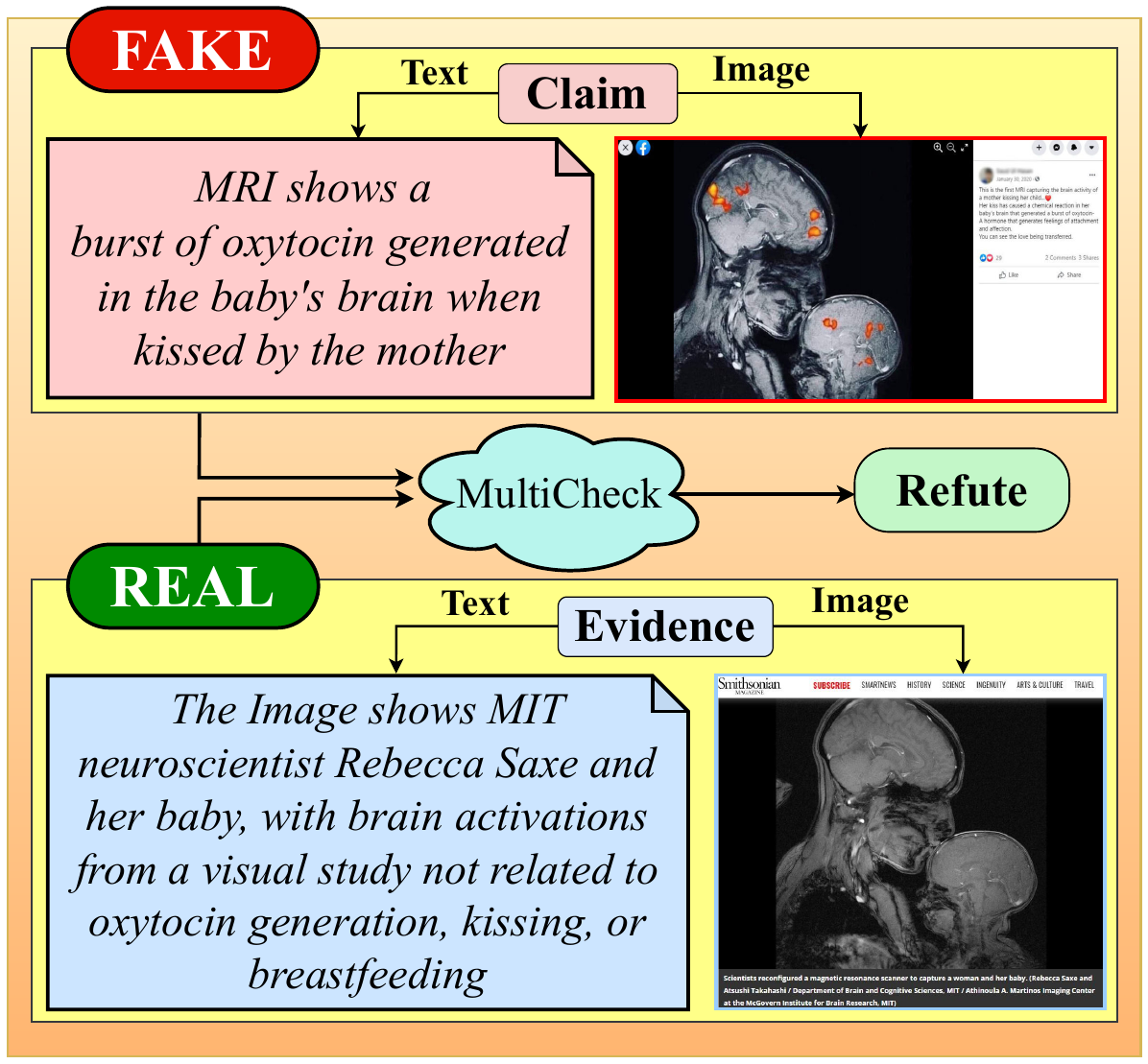}
\caption{Refuting a viral claim using combined text and image evidence.}
\Description{Refuting a viral claim using combined text and image evidence.}
\label{fig:motivation_ex}
\vspace{-2mm}
\end{figure}
\noindent As illustrated in Figure~\ref{fig:motivation_ex}, a viral claim may be paired with a scientific-looking image that appears credible but is actually unrelated. Detecting such inconsistencies requires going beyond surface-level alignment to capture nuanced cross-modal relationships. 

Despite this progress, multimodal fact-checking remains an open challenge. Unlike simple classification, it requires structured reasoning to determine whether modalities truly support or contradict each other. Existing approaches have generally followed one of the two strategies: (i) concatenating text–image embeddings \cite{chen2020uniter, lu2019vilbert}, or (ii) performing late fusion of modality-specific encoders \cite{du2023precofactv2}. For example, \textsc{Mocheg} \cite{yao2023mocheg} processes claims and evidence with separate encoders before shallow fusion, while \textsc{Pro-CoFactv2} \cite{du2023team} leverages attention-based mechanisms. Although these models capture imprecise correlations, they often miss fine-grained contradictions and struggle to understand modality-specific support signals, leading to limited transparency in reasoning about a claim. To overcome these limitations, we introduced "\textbf{MultiCheck}", a unified multimodal fact-checking framework that explicitly models cross-modal interactions. Our design integrates two key components: (i) a \textbf{fusion module} that captures semantic relations through element-wise difference and product operations, and (ii) a \textbf{contrastive learning objective} that aligns semantically consistent claim–document pairs within a shared latent space, thereby improving representation consistency across modalities. This approach is motivated by prior advances in multimodal relational reasoning, where element-wise operations have proven effective in capturing fine-grained interactions in natural language inference \cite{conneau2017supervised} and bilinear attention models \cite{kim2018bilinear}. These operations enable the model to encode both alignment and divergence between modalities, producing richer cross-modal representations. To further enhance semantic alignment, we introduced a contrastive head that applies a symmetric InfoNCE loss \cite{oord2018representation} on projected claim–document embeddings, pulling together related pairs while pushing apart unrelated ones. Unlike prior approaches that rely on frozen embeddings or shallow probing over large VLMs such as PaLI-Gemma \cite{cekinel-etal-2025-multimodal}, often wrapped in multi-stage agentic pipelines \cite{braun2024defame}, MultiCheck is fully trainable end-to-end. It uses compact text and image encoders, a simple relational fusion module based on element-wise difference and product, and a supervised contrastive head to jointly optimize classification and alignment. This purely discriminative design avoids heavy generative decoding and complex retrieval logic while still delivering competitive performance, making MultiCheck more suitable for resource-constrained, energy-aware multimodal fact verification. Our contributions are following,

\vspace{-1.0mm}

\begin{itemize}
    \item We proposed \textbf{MultiCheck}, a unified multimodal fact checking architecture that jointly models structured text, images, and OCR signals. We did an extensive experimental study across two benchmarks, Factify-2 and Mocheg, with multiple language vision backbones, fusion strategies, and training objectives. Our approach surpasses the baselines by a huge margin (gain of \textbf{47\%} and \textbf{33\%} in macro-f1 for factify2 and mocheg, respectively).

    \item We performed a comprehensive error analysis using statistical significance tests, showing that our model not only outperforms the baseline but does so in a structurally meaningful way, correcting more errors than it introduces. We supported our findings through qualitative analysis, highlighting how OCR cues and visual metadata can decisively shift predictions in subtle cases often missed by prior systems.

    \item Beyond standard evaluations, we further investigated MultiCheck’s robustness under modality imbalance, OCR noise, and contrastive loss hyperparameters ($\lambda$, $\tau$), providing quantitative evidence of stability and graceful degradation.
\end{itemize}

%% file: 20RelatedWorks.tex
\section{Related works:}
\label{related_work_1}
Recent research in automated fact-checking has highlighted the growing importance of incorporating multiple modalities, to tackle the diverse and evolving forms of misinformation \cite{abdelnabi2022open}. Early works, such as FEVER \cite{thorne2018fever} and CLEF2018 \cite{nakov2021automated} are primarily focused on verifying textual claims, laying foundational methods for claim verification based solely on textual evidence. However, later studies found that misinformation exploits images, videos, and audio alongside text to build convincing narratives \cite{hameleers2020picture,alam2022survey}. These studies have revealed the limitations of purely text-based fact-checking methods and sparked a shift toward multimodal fact-checking. To tackle the limitation of text-based methods, systems were designed to jointly process and reason over diverse types of content. For example, several studies, such as \cite{du2023precofactv2} and \cite{zlatkova-etal-2019-fact, khaliq-etal-2024-ragar}, have explored architectures that integrated textual and visual features. These models employ mechanisms like attention or contrastive learning to enhance detection accuracy. Recent work has also leveraged large vision–language models (VLMs) for fact verification. \citet{cekinel-etal-2025-multimodal} proposed a probing-based classifier that uses a frozen PaLI-Gemma VLM with lightweight heads, achieving strong results on Factify-2. \cite{braun2024defame} introduced DEFAME, a dynamic agent that combines multiple multimodal experts through a multi-stage retrieval and reasoning pipeline. These approaches showcase the power of VLMs but incur high computational and energy costs and require complex inference pipelines. In addition to these developments, comprehensive surveys, such as \cite{akhtar2023multimodal}, offer a detailed overview regarding the emerging field of multimodal fact-checking. They highlighted both the technical challenges and the promising research directions ahead.
Key challenges include aligning information across modalities, managing incomplete or noisy evidence, and ensuring scalability for practical deployment. Despite significant progress, effectively integrating multimodal information remains an open research problem. This challenge continues to motivate the development of new architectures and learning methods. Robust fact verification in multimodal contexts still requires innovative solutions.

%% file: 30Dataset.tex
\section{Dataset details:}
\label{dataset_details}

\begin{figure}[ht]
\centering
\includegraphics[width=\columnwidth]{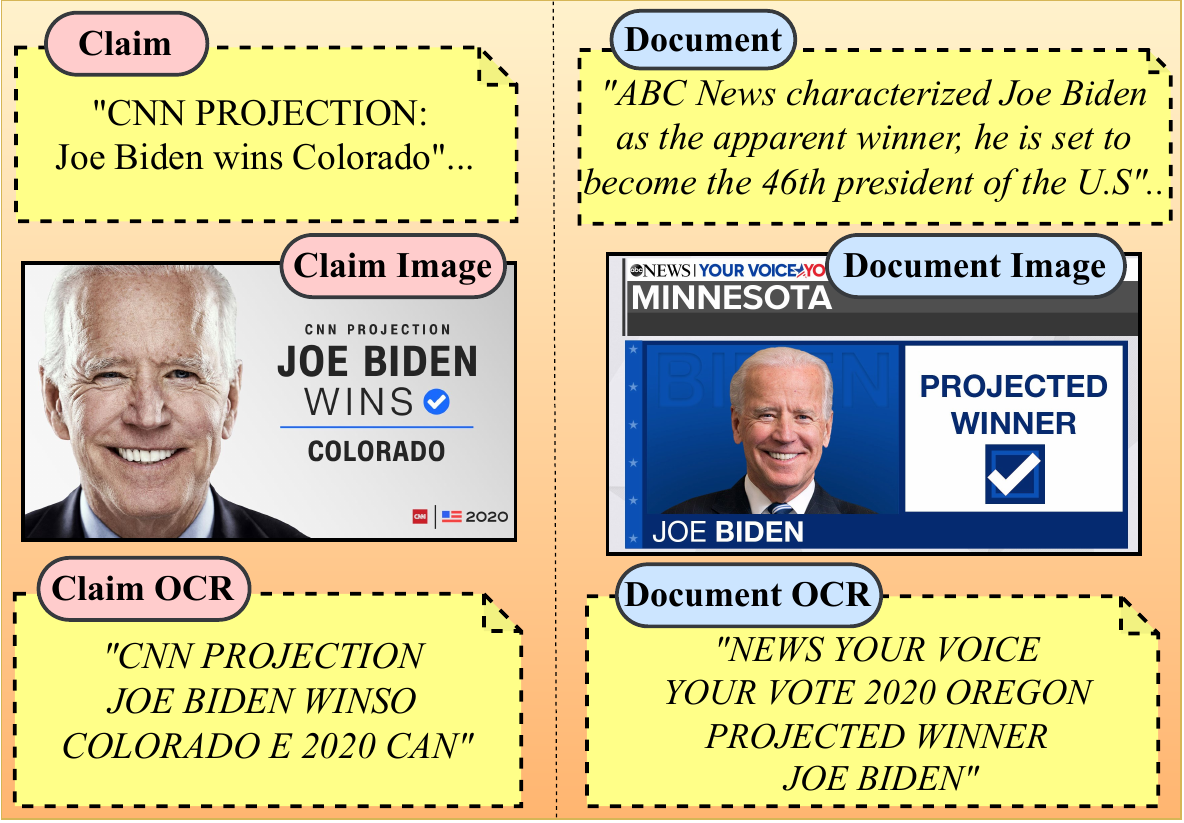}
\caption{Example of a sample from the Factify 2.}
\Description{Example of a sample from the Factify 2.}
\label{fig:Dataset_sample}
\end{figure}

For our experiments, we use two multimodal fact-checking datasets: Factify-2 \cite{suryavardan2022factify} and Mocheg \cite{yao2023end}. Factify-2 contains 50,000 claim– evidence pairs sourced from verified news outlets and fact-checking platforms. Each instance includes the claim text, supporting or refuting evidence text, an associated image, and OCR-extracted text isllustrated in Figure \ref{fig:Dataset_sample}, and is annotated using a five-class labeling scheme. Since the official test set is not publicly available, we follow the protocol of \citet{cekinel-etal-2025-multimodal} and treat the validation set as test data, reallocating 7,500 training examples to form a new validation split. Mocheg, in contrast, comprises 15,601 claims collected from fact-checking websites\footnote{\url{https://www.politifact.com}} and annotated with three labels. The dataset includes a large amount of retrieved multimodal evidence, consisting of text and images taken from journalist-verified rulings. This makes Mocheg a stronger and more challenging open-domain benchmark. We used the official train/validation/test split provided in our experiments. The distribution of samples across classes for both datasets are reported in Table~\ref{tab:dataset_stats}.


\input{tables/Dataset_stats}

\input{tables/Dataset_stats_2}

%% file: tables/Dataset_stats.tex
\begin{table}[ht]
\centering
\setlength{\tabcolsep}{6pt}
\renewcommand{\arraystretch}{1.15}

\resizebox{0.48\textwidth}{!}{
\begin{tabular}{l|r|r|r|r}
\toprule[0.12em]
\textbf{Classes} & \textbf{Train} & \textbf{Validation} & \textbf{Test} & \textbf{Total} \\
\midrule
\multicolumn{5}{c}{\textbf{Factify 2 Dataset} \cite{suryavardan2022factify}} \\
\midrule
Support\_Multimodal & 5,580 & 1,420 & 1,500 & 8,500 \\
Support\_Text & 5,485 & 1,515 & 1,500 & 8,500 \\
Insufficient\_Multimodal & 5,472 & 1,528 & 1,500 & 8,500 \\
Insufficient\_Text & 5,494 & 1,506 & 1,500 & 8,500 \\
Refute & 5,469 & 1,531 & 1,500 & 8,500 \\
\midrule
\textbf{Total (Factify 2)} & 27,500 & 7,500 & 7,500 & 42,500 \\
\midrule
\multicolumn{5}{c}{\textbf{Mocheg Dataset} \cite{yao2023mocheg}} \\
\midrule
Supported & 3,826 & 501 & 817 & 5,144 \\
Refuted   & 4,542 & 488 & 825 & 5,855 \\
Not Enough Info & 3,301 & 501 & 800 & 4,602 \\
\midrule
\textbf{Total (Mocheg)} & 11,669 & 1,490 & 2,442 & 15,601 \\
\bottomrule[0.12em]
\end{tabular}}
\caption{Dataset statistics for Factify 2 and Mocheg across training, validation, and test splits.}
\label{tab:dataset_stats}
\end{table}



%% file: tables/Dataset_stats_2.tex


%% file: 40Methods.tex
\section{Methodology:}
\label{sec:methodology}

\begin{figure*}[ht]
    \centering
    \includegraphics[width=1.00\linewidth]{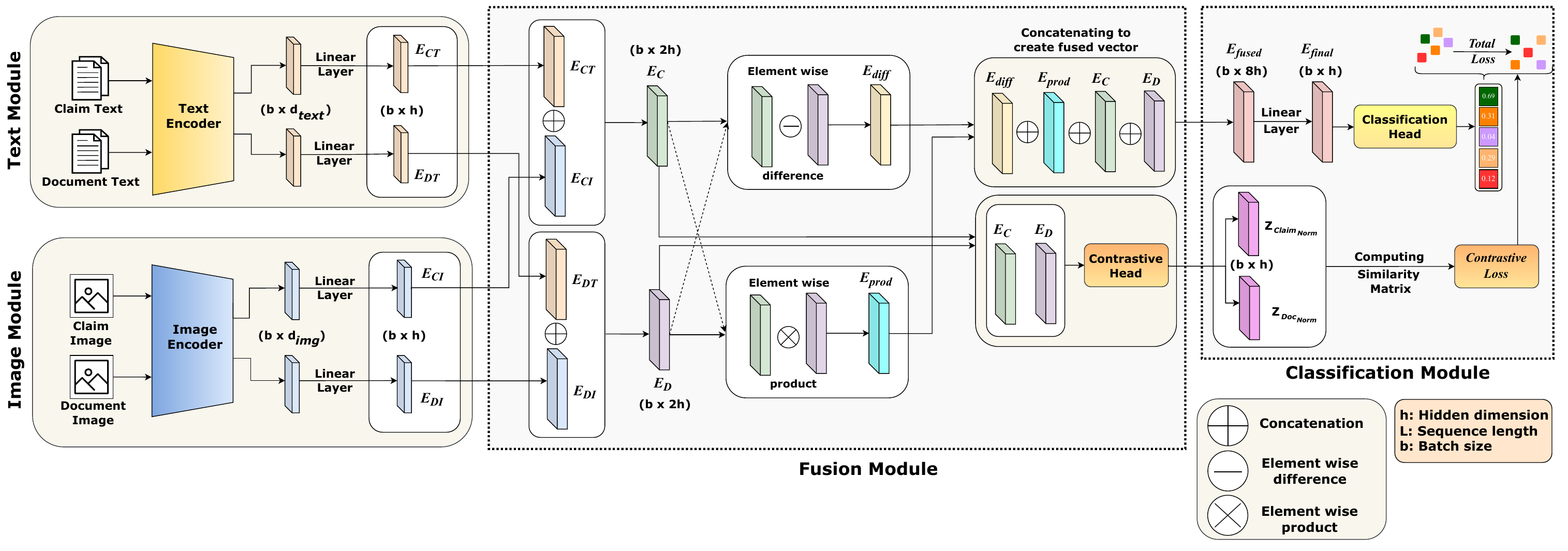}
    \caption{Intuitive fusion representation using element-wise difference and product.}
    \label{fig:intuitive-fusion}
\end{figure*}

In this section, we present our proposed framework, illustrated in Figure \ref{fig:intuitive-fusion}. The framework consists of four components: (i) a text module, (ii) an image module, (iii) a fusion module, and (iv) a classification module. We describe each component in detail below.

\begin{figure}[ht]
\centering
\includegraphics[width=\columnwidth]{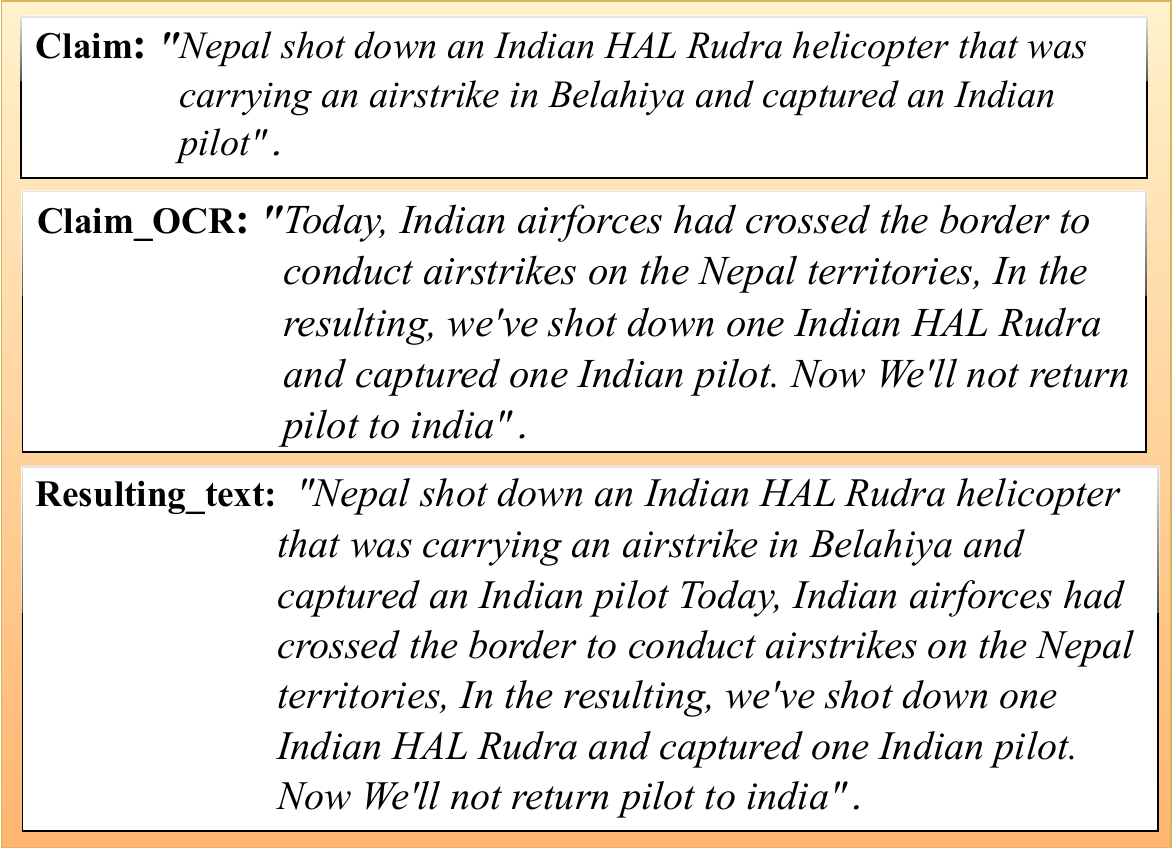}
\caption{\textit{Illustration:} The claim suggests Nepal shot down an Indian helicopter. However, the OCR text contradicts this by suggesting Indian aggression, not Nepali. Without OCR, the model could misclassify this. The fused text representation enables correct "Refute" labeling.}
\label{fig:text_concat_photo}
\end{figure}

\noindent \textbf{Text module:}
\label{sec: text module main} 
We used this module to generate text embeddings for claims and evidence documents. To create a claim embedding, we first concatenated the claim text with the OCR-extracted text from the claim image and then passed the combined input to a text encoder. We followed the same procedure to create the document embeddings (see Figure \ref{fig:text_concat_photo}). We employed pre-trained language models such as RoBERTa \cite{liu2019roberta}, DeBERTa \cite{he2020deberta}, and SBERT \cite{reimers2019sentence} as text encoders. Finally, we used a linear layer to ensure uniformity between the text and image module embeddings. Mathematically,
 
\vspace{-2mm}
\begin{align*}
{E}_{\text{\textit{CT }}} \in \mathbb{R}^{b \times h }\\
{E}_{\text{\textit{DT}}} \in \mathbb{R}^{b \times h}
\end{align*}

\noindent Where the claim and evidence document text embeddings are represented by $E_{\text{\textit{CT}}}$ and $E_{\text{\textit{DT}}}$, respectively. Further,
$h$ and $b$ represent the common latent space dimensionality and the batch size, respectively.

\noindent \textbf{Image module:}
\label{sec: image main module} 
We use the image module to obtain embeddings for the images associated with claims and their corresponding evidence documents. To create an image embedding, we pass each image through a vision encoder, either ResNet50~\cite{he2016deep} or the Vision Transformer (ViT)~\cite{dosovitskiy2020image}. We then apply a linear layer to ensure uniformity between the image and text module embeddings. Mathematically,


\vspace{-1.5mm}
\begin{align*}
    {E}_{\text{\textit{CI }}} \in \mathbb{R}^{b \times h}\\
    {E}_{\text{\textit{DI}}} \in \mathbb{R}^{b \times h}
\end{align*}

\noindent Where the claim and evidence document image embeddings are represented by $E_{\text{\textit{CI}}}$ and $E_{\text{\textit{DI}}}$, respectively. Similarly, $h$ and $b$ represent latent space dimensionality and the batch size like previously.  

\noindent \textbf{Fusion module:}
\label{sec: fusion module main} 
Our key innovation lies in the fusion module. While prior systems relied on simple feature concatenation \cite{mishra2020generating, sata2025multimodal, wang2022ita}, our framework employed element-wise difference and product between claim and evidence embeddings to explicitly capture their alignment and divergence. Our approach is inspired by past works \cite{kim2018bilinear, conneau2017supervised, chen2020uniter, liu2023cross, gong2024heterogeneous}, where a similar philosophy is used for natural language inference and modeling relational reasoning across modalities. We took ${E}_{\text{\textit{CT }}}, {E}_{\text{\textit{DT }}} , {E}_{\text{\textit{CI }}}$ and, ${E}_{\text{\textit{DI }}}$ generated by prior modules as input. The multimodal representation embeddings of claims and evidence documents were generated by concatenating text and image embeddings. Mathematically, 

\vspace{-3mm}
\begin{align*}
{E}_{\text{\textit{C}}} &= {E}_{\text{\textit{CT}}} \oplus {E}_{\text{\textit{CI}}}, \text{where } {E}_{\text{\textit{C}}}\in \mathbb{R}^{b \times 2h} \\
{E}_{\text{\textit{D}}} &= {E}_{\text{\textit{DT}}} \oplus {E}_{\text{\textit{DI}}}, \text{where } {E}_{\text{\textit{D}}}\in \mathbb{R}^{b \times 2h}
\end{align*} 

\noindent Where $\oplus$ represents the concatenation operation. We then computed \textit{element-wise difference} and \textit{product} to highlight the differences and alignment between ${E}_{\text{\textit{C}}}$ and ${E}_{\text{\textit{D}}}$. Past works \cite{conneau2017supervised, kim2018bilinear} show that these vectors offer a feature-level map of alignment and mismatch between the text and the image, helping the model identify key agreements and contradictions across them \cite{zhang2024unsupervised}. Mathematically, 

\vspace{-5mm}
\begin{align*}
{E}_{\text{diff}} &= \left| {E}_{\text{\textit{C}}} \ominus {E}_{\text{\textit{D}}} \right| ,   \text{where } {E}_{\text{diff}}\in \mathbb{R}^{b \times 2h}\\
{E}_{\text{prod}} &= {E}_{\text{\textit{C}}} \otimes {E}_{\text{\textit{D }}} ,   \text{where } {E}_{\text{prod}}\in \mathbb{R}^{b \times 2h}
\end{align*}
where $\ominus$ and $\otimes$ denote element-wise difference and product, respectively. Finally, we constructed a fused embedding by concatenating $ {E}_{\text{diff}}, {E}_{\text{prod}}, {E}_{\text{\textit{C }}}, \text{and } {E}_{\text{\textit{D }}}$. Mathematically,
\vspace{-1.5mm}
\begin{align*}
    E_{\text{fused}} = E_{\text{diff}} \oplus E_{\text{prod}} \oplus {E}_{\text{C}} \oplus {E}_{\text{D}},   \text{where } {E}_{\text{fused}}\in \mathbb{R}^{b \times 8h}
\end{align*}  

\noindent We believe this fused embedding robustly captures the refined interactions between text and image, providing a compact representation well-suited for veracity classification. Simultaneously, we integrated a contrastive projection head \cite{wang2020understanding} that encourages semantic alignment between matching claim-document pairs. Specifically, the embeddings ${E}_{\text{\textit{C}}} \text{ and } {E}_{\text{\textit{D}}}$ are passed through a shared two-layer projection network to obtain projected embeddings. Mathematically,
\vspace{-1mm}
\begin{align*}
\mathbf{Z}_{\text{claim}} &= f_{\text{proj}}({E}_{\text{\textit{C}}}) \\
\mathbf{Z}_{\text{doc}} &= f_{\text{proj}}({E}_{\text{\textit{D}}})
\end{align*}
where the projection function $f_{\text{proj}}$ is defined as:
\begin{align*}
f_{\text{proj}}({\textit{E}}) = \mathbf{W}_2 \cdot \text{ReLU}({W}_1 \cdot {E} + \mathbf{B}_1) + \mathbf{B}_2
\end{align*}
\vspace{-5mm}
\begin{align*}
\text{where}, \mathbf{W}_1 \in \mathbb{R}^{2h\times h} ,\mathbf{W}_2 \in \mathbb{R}^{h\times h}, \text{and }   \mathbf{B}_1,\mathbf{B}_2 \in \mathbb{R}^{h}
\end{align*}

\noindent where, $\mathbf{W_1}$, $\mathbf{W_2}$, $\mathbf{B_1}$ and $\mathbf{B_2}$ are weights and biases. This bottleneck structure compresses the dimension from $2\textbf{\textit{h}}$ to $\textbf{\textit{h}}$ and reprojects it, yielding richer discriminative embeddings. Reprojecting these vectors from $2h$ to $h$ removes duplication from concatenation, compressing features into a compact, discriminative space \cite{chen2020simple}. It balances contributions from text and image modalities, improving cross-modal alignment \cite{radford2021learning}. This shared latent space ensures cosine similarity captures true semantic claim–document alignment rather than raw backbone artifacts \cite{oord2018representation}. The resulting vector embeddings, $\mathbf{Z}_{\text{claim}}$ and $\mathbf{Z}_{\text{doc}}$, are used to compute a symmetric InfoNCE loss~\cite{oord2018representation} during training. This loss pulls embeddings of matching claim–document pairs closer and pushes unrelated ones apart. By enforcing semantic alignment across modalities, it sharpens the model’s ability to capture subtle cross-modal differences, improving both generalization and robustness.

\noindent \textbf{Classification module:}
We used this module to classify the category of a claim over fine-grained veracity labels. We take the fused embedding vector from the fusion module and process it through a feedforward network with GELU activation \cite{hendrycks2016gaussian}, and dropout regularization. The resulting representation serves as the final embedding vector for classification. It is then passed to the classification head, which outputs predictions across veracity labels. Mathematically,
\vspace{-1.5mm}
\begin{align*}
{E}_{\text{final}} = \text{FFN}({E}_{\text{fused}}) \in \mathbb{R}^{b \times h}
\end{align*}

\noindent This final vector ${E}_{\text{final}}$ is passed through a linear classifier:
\begin{align*}
\mathbf{P} = \mathbf{W}_{\text{cls}} \cdot {E}_{\text{final}} + \mathbf{B}, \quad \mathbf{P} \in \mathbb{R}^{b \times k}
\end{align*}
where $\mathbf{P}$ contains the predicted logits over the $k$ veracity labels. We used the standard cross-entropy loss to supervise this classification. Mathematically,
\[
\mathcal{L}_{\text{CE}} 
=
-\frac{1}{b}
\sum_{i=1}^{b}
\log
\left(
\frac{
\exp\left( Z_{i, y_i} \right)
}{
\sum_{j=1}^{k} \exp\left( Z_{i,j} \right)
}
\right)
\]
Here, $\mathbf{Z} \in \mathbb{R}^{b \times k}$ is the logit matrix, where $Z_{i,j}$ is the logit for sample $i$ and class $j$, and $y_i$ is the index of the correct class for sample $i$.
Simultaneously, to improve discriminative learning and cross-modal alignment, we employed a symmetric contrastive loss (InfoNCE). The projected embeddings $\mathbf{Z}_{\text{claim}}$ and $\mathbf{Z}_{\text{doc}}$ are normalized to unit length following L2 Norm, ensuring that similarity comparisons depend only on their semantic content\cite{oord2018representation}. Mathematically,
\begin{align*}
\hat{\mathbf{z}} = \frac{\mathbf{z}}{\lVert \mathbf{z} \rVert_2}
\end{align*}
We then computed the pairwise similarity matrix $\mathbf{S}$ as:
\[\mathbf{S} = \hat{\mathbf{Z}}_{\text{claim}} \cdot \hat{\mathbf{Z}}_{\text{doc}}^\top \in \mathbb{R}^{b \times b}\]
The $S_{ij}$ represents the cosine similarity between the $i$-th claim and $j$-th document. Then the InfoNCE loss is computed in both directions: (i) \textbf{claim $\rightarrow$ evidence document}  and (ii) \textbf{document evidence $\rightarrow$ claim}

\[
\text{Loss}_{\text{claim} \rightarrow \text{doc}} 
=
-\frac{1}{b}
\sum_{i=1}^{b}
\log
\frac{
\exp\left(S_{ii} / \tau \right)
}{
\sum_{j=1}^{b} \exp\left(S_{ij} / \tau \right)
}\]
\[
\text{Loss}_{\text{doc} \rightarrow \text{claim}} 
=
-\frac{1}{b}
\sum_{j=1}^{b}
\log
\frac{
\exp\left(S_{jj} / \tau \right)
}{
\sum_{i=1}^{b} \exp\left(S_{ij} / \tau \right)
}
\]
Where, $b$ is the batch size and $\tau>0$ is a temperature
hyperparameter that controls the sharpness of the softmax. In $\mathcal{L}_{\text{claim}\rightarrow\text{doc}}$,
each row $i$ treats document $i$ as the positive example and all
other documents where $j \neq i$ in the batch as in-batch negatives:
the numerator $\exp(S_{ii}/\tau)$ is the score of the true
claim–document pair, while the denominator
$\sum_{j=1}^{b} \exp(S_{ij}/\tau)$ sums over all documents in
the batch. The reverse loss
$\mathcal{L}_{\text{doc}\rightarrow\text{claim}}$ is defined
analogously by viewing each document as the query and each claim
as a candidate match. Then we take the average of both to calculate the final contrastive loss:
\[
\mathcal{L}_{\text{contrastive}}
=
\frac{1}{2}
\left(
\text{Loss}_{\text{claim} \rightarrow \text{doc}}
+
\text{Loss}_{\text{doc} \rightarrow \text{claim}}
\right)
\]
The total training objective combines classification and contrastive supervision:
\begin{align*}
\mathcal{L}_{\text{total}} = \mathcal{L}_{\text{CE}} + \lambda \cdot \mathcal{L}_{\text{contrastive}}
\end{align*}
Where $\lambda$ balances the influence of the contrastive signal. We configured $\lambda = 0.1$ in all experiments.  We believe this dual-objective training not only improves classification accuracy but also structures the latent space such that semantically aligned claim-document pairs are closer, while unrelated pairs remain well-separated.  Empirically, the inclusion of the contrastive head consistently yields performance gains across all evaluated models. In line with prior findings, InfoNCE-based contrastive learning enhances discriminative representations across modalities \cite{oord2018representation, radford2021learning}.

%% file: 50Experiments.tex
\section{Experiments:}
\noindent In this section, we have reported our experiments to evaluate the effectiveness of the proposed framework.
\vspace{-3mm}
\subsection{Experimental setup:}

\noindent All experiments were conducted on a single NVIDIA A100 80GB GPU using PyTorch and Hugging Face Transformers. We evaluated the model performance using the macro F1 score, which balances class-wise F1 scores based on support (i.e., true instance count per class). We performed thorough hyperparameter tuning for both the contrastive and non-contrastive variants. The final configurations for the experiments are reported in Table \ref{tab:hyperparams}. We assessed how (i) learning strategies, (ii) language and vision models, and (iii) fusion methods with element-wise operations affect the performance.

\input{tables/hyperparameter}


\vspace{-3mm}
\subsection{Baselines:}
\label{section:baselines}

For comparative evaluation, we considered two state-of-the-art models as baselines. They are explained in the following, 

\noindent \textbf{Baseline 1:}  We reproduced the \textbf{Pro-CoFactv2} model, originally proposed by~\cite{du2023team} as our first baseline. This model leverages large pretrained encoders: DeBERTa-large for text and SwinV2-base for images, along with co-attention blocks, adapter layers, and a feature extractor. Both encoders are unfrozen and trained end-to-end. It applies linear fusion of text and image features, and jointly optimises a multi-loss objective: cross-entropy for classification and a supervised contrastive loss scaled by 0.3 to improve intra-class alignment in the embedding space. We replicated the original configuration using the same random seeds (42, 57, 196, 906), learning rate ($5 \times 10^{-5}$), batch size (32), and training duration (20 epochs). Sequence length is set to 128 due to GPU constraints, with a dropout rate of 0.1, 12 attention heads, and an intermediate hidden dimension of 256. Reproduced baseline performance is reported in Table~\ref{tab:baseline_performance}.

\noindent \textbf{Baseline 2:} We also compared MultiCheck against the PaLI-Gemma-based system of \citet{cekinel-etal-2025-multimodal}. Their work investigates multimodal fact-checking with large vision language models and reports both probing and fine-tuning results on Mocheg and Factify-2.   For a fair comparison with our fully supervised setting, we use their \emph{fine-tuned} PaLI-Gemma-3B model as a strong VLM baseline. Following \citet{cekinel-etal-2025-multimodal}, we adopted the three-way label mapping on Factify-2 (merging \textit{Support\_Text} / \textit{Support-Multimodal} into \textit{Support}, and \textit{Insufficient\_Text} / \textit{Insufficient-Multimodal} into \textit{NEI}) and use the same three-way scheme on Mocheg. We directly compared our macro F1 scores on these three-way setups to the reported PaLI-Gemma fine-tuning results (Factify-2: 0.83 macro F1; Mocheg: 0.36 macro F1), treating them as a high-capacity, VLM-based reference point for end-to-end training rather than zero-shot prompting or pure probing. Although, due to our resource constraints, we were unable (it requires 60GB GPU memory and 37 hours to complete an epoch) to reproduce this baseline, we are reporting the actual values claimed by it for comparison to MultiCheck.


\input{tables/Baseline}

\vspace{-2.5mm}
\subsection{Experimental variants:}

To thoroughly evaluate our models, we considered three variants of them. They are, 

\begin{itemize}
    \item \textbf{With contrastive Head}: This version integrates the contrastive projection head applied to ${E}_{\text{\textit{C}}} \text{ and } {E}_{\text{\textit{D}}}$. Training includes a contrastive loss in addition to the standard cross-entropy loss. We believe this encourages the model to learn modality-consistent and semantically aligned embeddings.

    \item \textbf{With-out contrastive Head}: In this version, the contrastive projection head and the associated loss are omitted. The model relies solely on cross-entropy loss for classification. 

    \item \textbf{Quantised version}: To study efficient resource-conscious configurations, we additionally considered the quantized version of LLMs with 4-bit QLoRA \cite{dettmers2023qlora} adapters in MultiCheck. Specifically, we considered (i) the 4-bit quantized versions of the considered language models (DeBERTa, RoBERTa and SBERT), (ii) 4-bit quantized versions of LLaMA-3.1-8B and Mistral-7B with 4-bit QLoRA adapters. They are trained end-to-end with the image encoder, while the contrastive head configurations remain unchanged (see Section 9.1 in the Appendix).
    
\end{itemize}







%% file: tables/hyperparameter.tex
\begin{table}[htbp]
  \centering
  \scriptsize
  \setlength{\tabcolsep}{6pt}
  \renewcommand{\arraystretch}{1.2}
  \begin{tabular}{@{} >{\bfseries}l l l @{}}
    \toprule
    Component            & Hyperparameter          & Value \\
    \midrule

    Reproducibility      & Seeds                   & 42, 57, 196, 906 \\
    Tokenization         & Max length              & 128 \\

    Contrastive Loss     & Temperature ($\tau$)             & 0.1 \\
                         & Loss weight ($\lambda$) & 0.1 \\

    Optimization         & Optimizer               & Adam \\
                         & Learning rate           & $1\times10^{-5}$ \\
                         & Batch size              & 32 \\
                         & Num workers             & 4 \\
                         & Epochs                  & 20 \\

    LR Scheduling        & Scheduler               & ReduceLROnPlateau \\

    Early Stopping       & Patience                & 5 \\
    \bottomrule
  \end{tabular}
  \caption{Hyperparameter settings used in all experiments.}
  \label{tab:hyperparams}
\end{table}

%% file: tables/Baseline.tex
\newcommand{\unc}[1]{{\footnotesize(\textpm#1)}}

\begin{table}[htbp]
\centering
\setkeys{Gin}{keepaspectratio}
\resizebox{0.48\textwidth}{!}{%
\begin{tabular}{@{}l|c|c|c@{}}
\toprule[0.12em]
\textbf{Class} & \textbf{Precision} & \textbf{Recall} & \textbf{F1 Score} \\
\midrule
Support Text            & 0.48 (\textpm 0.04) & 0.38 (\textpm 0.06) & 0.42 (\textpm 0.03) \\
Support Multimodal      & 0.50 (\textpm 0.04) & 0.61 (\textpm 0.02) & 0.55 (\textpm 0.02) \\
Insufficient Text       & 0.50 (\textpm 0.01) & 0.44 (\textpm 0.07) & 0.46 (\textpm 0.04) \\
Insufficient Multimodal & 0.43 (\textpm 0.02) & 0.46 (\textpm 0.06) & 0.44 (\textpm 0.02) \\
Refute                   & 0.98 (\textpm 0.00) & 0.98 (\textpm 0.00) & 0.98 (\textpm 0.00) \\
\midrule
\textbf{Macro F1 Score} & \multicolumn{3}{|c}{\textbf{0.57} (\textpm 0.01)} \\
\bottomrule[0.12em]
\end{tabular}
}
\caption{Performance of the reproduced baseline 1 model (Pro-CoFactv2) on the Factify 2 dataset.}
\label{tab:baseline_performance}
\end{table}

%% file: 60Results.tex
\section{Results and Discussion:}
\label{Results_and_discussion}

\input{tables/results_2}
\input{tables/results_1}

\input{tables/Mocheg_results_Cont}

\input{tables/mocheg_QLoRA}
\input{tables/mocheg_result_no_contrastive_head}
\input{tables/factify_three_way}

\input{tables/factify_three_way_noncont}

\noindent In the following, we have reported our findings for various experimental setups. For Mocheg, we have reported results for the original 3-class labels. For Factify-2, we have reported results for both the 5-class and 3-class variants, allowing for direct comparison with different baselines. Some of the tables are kept in the appendix due to space constraints.

\subsection{With contrastive head:}
\begin{itemize}
    \item \textbf{Factify2/5-class:- } For this configuration, we have reported the results in Table \ref{tab:factify2_results}. Many of our model variants, such as RoBERTa+ResNet50, RoBERTa+ViT, and  DeBERTa+ViT having the highest performance($\sim MF1:\textbf{0.84}, \textbf{47}\%\uparrow$), outperform the highest macro-f1 reported baseline ($MF1: 0.57$:, see Table \ref{tab:baseline_performance}). Gains are particularly high for the “Insufficient” ($\textbf{78}\%\uparrow$ gain) and “Support” ($\textbf{68}\%\uparrow$ gain) categories. All improvements are statistically significant (see, Table \ref{tab:significance_contrastive_tests}).  
    
    
    \item \textbf{Factify2/3-class:- } The results for this configuration are reported in Table \ref{tab:factify_three_way_label}. Several MultiCheck variants, such as RoBERTa+ResNet50, RoBERTa+ViT, DeBERTa+ViT, DeBERTa +ResNet50, and SBERT+ResNet50, outperform PaLI-Gemma (baseline 2) by a large margin (a maximum of $\sim MF1:\textbf{0.90}, \textbf{91}\%\uparrow$). The largest gains are observed for ``Support" (~$\textbf{15}\%\uparrow$) and ``NEI" (~$\textbf{15}\%\uparrow$), while ``Refute" remains essentially saturated ($\sim \textbf{1.00}$).   
    
    \item \textbf{Mocheg/3-class:- }: We have reported the results of this configuration in Table \ref{tab:mocheg_results}. Several MultiCheck variants, such as RoBERTa+ResNet50, DeBERTa+ResNet50 and RoBERTa+ViT, outperform PaLI-Gemma (baseline 2) by a significant margin (maximum $\sim MF1:\textbf{0.48}, \textbf{33}\%\uparrow$)). All the labels, ``Support" (~$\textbf{24}\%\uparrow$), ``NEI" (~$\textbf{117}\%\uparrow$), and ``Refute" (~$\textbf{29}\%\uparrow$) showed substantial improvements.

\end{itemize}

\subsection{Without contrastive head:}
\begin{itemize}
     \item \textbf{Factify2/5-class:-} For this configuration, we have reported the results in Table \ref{tab:factify2_results_no_contrastive}. Many of our model variants, such as Roberta + ResNet50, Roberta + ViT and DeBERTa + ViT outperform the highest macro-f1 reported baseline ($MF1$: 57, refer Table 15). All three of our best models achieved a macro-f1 score of 0.82 with a gain of ($\sim \textbf{43.85}\%\uparrow$) compared to the baseline. Gains are particularly significant for the “Insufficient” and “Support” categories. All improvements are statistically significant (see, Table 14)
    \item \textbf{Factify2/3-class:-} The results for this configuration are reported in Table \ref{tab:factify_three_way_label_non_cont}. Several MultiCheck variants, such as Roberta + ResNet50 ($\sim MF1:\textbf{0.86}, \textbf{2.38}\%\uparrow$) , Roberta + ViT ($\sim MF1:\textbf{0.87}, \textbf{3.57}\%\uparrow$), DeBERTa + ViT ($\sim MF1:\textbf{0.87}, \textbf{3.57}\%\uparrow$), DeBERTa + ResNet50 ($\sim MF1:\textbf{0.88}, \textbf{4.76}\%\uparrow$) and SBERT + ResNet50 ($\sim MF1:\textbf{0.86}, \textbf{2.38}\%\uparrow$) outperform PaLI-Gemma (baseline 2) ($ MF1:\textbf{0.84}$). The largest gains are observed for ``Support" (~$\%$) and ``NEI" (\textbf{0.76→0.85}), while ``Refute" remains essentially saturated (~$MF: 1$).
    \item \textbf{Mocheg/3-class:-}: We have reported the results of this configuration in Table \ref{tab:mocheg_results_without_contrastive_head}. Several MultiCheck variants, such as Roberta + ResNet50 ($\sim MF1:\textbf{0.43}, \textbf{19.4}\%\uparrow$), and DeBERTa + ResNet50 ($\sim MF1:\textbf{0.40}, \textbf{11.1}\%\uparrow$), outperform PaLI-Gemma (baseline 2) by a significant margin.
\end{itemize}

\subsection{Quantized Version:}

As we didn't see much variation in performance upon changing the random seeds, for quantized experiments, we produced results for a single seed.  Our observatons are following,

\begin{itemize}
    \item \textbf{Factify2/5-class:-}  For this configuration, we have reported the results in Table \ref{tab:factify2_results_qlora}. Two models LLaMa-3.1-8B \textsubscript{QLoRA} + ResNet50 and Mistral7B \textsubscript{QLoRA} + ResNet50 achieved the highest macro-f1 performance of 0.84, which is equal to the macro-f1 performance of Roberta+ResNet50, Roberta+ViT and DeBERTa+ViT model for Factify 2 dataset. There is a drop in macro-f1 performance in quantized model like Roberta \textsubscript{QLoRA} + ResNet50 (MF1: 0.82) as compared to Roberta + ResNet50 (MF1: 0.84) by 2.38$\%\downarrow$. 
    \item \textbf{Mocheg/3-class:- }: We have reported results for this configuration in Table \ref{tab:mocheg_results_qlora}. Mistral7B \textsubscript{QLoRA} + ResNet50 model reported the highest macro-f1 score of 0.49 among all other quantized model for Mocheg dataset. It can be observed that there is a drop in macro-f1 performance for quantized model as compared to model with contrastive head for this dataset as well. Interestingly, the DeBERTa \textsubscript{QLoRA} + ViT has achieved macro-f1 value of 0.47 which is higher than the macro-f1 value of DeBERTa + ViT model (MF1: 0.39) by 20.50$\%\uparrow$.
\end{itemize}

\subsection{Statistical significance test:} We did statistical significance tests with respect to baseline-1 (for the 5-class variant of Factify2). However, we could not do the same for baseline-2 (impacting the 3-class variant of Factify2 and Mocheg), which is not reproducible under a realistic GPU budget (requires 37 hours for a single epoch).  We applied McNemar’s test for overall accuracy and Bowker’s test for class-level shifts. Tables~\ref{tab:significance_contrastive_tests} show consistently high $\chi^2$ scores with p-values < 0.05, confirming significant improvements. Whereas Bowker’s results further indicate structured, non-random gains in class predictions.

%% file: tables/results_1.tex
\begin{table*}[htbp]
\centering
\setlength{\tabcolsep}{6pt}
\renewcommand{\arraystretch}{1.2}

\resizebox{\linewidth}{!}{
\begin{tabular}{l|c|c|c|c|c|>{\columncolor{gray!15}}c}
\toprule[0.12em]
\textbf{Models With Contrastive Head} &
\textbf{Support Text} &
\textbf{Support Multimodal} &
\textbf{Insufficient Text} &
\textbf{Insufficient Multimodal} &
\textbf{Refute} &
\textbf{Macro F1} \\
\midrule

Roberta + ResNet50
  & \bfseries0.77\unc{0.02}
  & 0.83\unc{0.01}
  & \bfseries0.82\unc{0.00}
  & 0.76\unc{0.02}
  & 1.00\unc{0.00}
  & \bfseries0.84\unc{0.01} \\

Roberta + ViT
  & \bfseries0.77\unc{0.02}
  & \bfseries0.84\unc{0.01}
  & \bfseries0.82\unc{0.00}
  & \bfseries0.77\unc{0.01}
  & 1.00\unc{0.00}
  & \bfseries0.84\unc{0.01} \\

DeBERTa + ViT
  & 0.77\unc{0.01}
  & 0.84\unc{0.01}
  & 0.83\unc{0.00}
  & 0.78\unc{0.01}
  & 1.00\unc{0.00}
  & \bfseries0.84\unc{0.01} \\

DeBERTa + ResNet50
  & 0.76\unc{0.02}
  & 0.81\unc{0.02}
  & 0.81\unc{0.01}
  & 0.75\unc{0.02}
  & 1.00\unc{0.00}
  & 0.83\unc{0.01} \\

SBERT + ResNet50
  & 0.75\unc{0.01}
  & \bfseries0.83\unc{0.01}
  & 0.79\unc{0.00}
  & \bfseries0.76\unc{0.01}
  & 1.00\unc{0.00}
  & 0.82\unc{0.00} \\

Baseline 1
  & 0.42\unc{0.03}
  & 0.55\unc{0.02}
  & 0.46\unc{0.04}
  & 0.44\unc{0.02}
  & 0.98\unc{0.00}
  & 0.57\unc{0.01} \\

\bottomrule[0.12em]
\end{tabular}
}
\caption{Class‐wise F1 scores and Macro F1 for model combinations \textbf{with contrastive head} on Factify 2.}
\label{tab:factify2_results}
\end{table*}

\begin{table*}[htbp]
\centering
\setlength{\tabcolsep}{6pt}
\renewcommand{\arraystretch}{1.2}

\resizebox{\linewidth}{!}{
\begin{tabular}{l|c|c|c|c|c|>{\columncolor{gray!15}}c}
\toprule[0.12em]
\textbf{QLoRA Models With Contrastive Head } &
\textbf{Support Text} &
\textbf{Support Multimodal} &
\textbf{Insufficient Text} &
\textbf{Insufficient Multimodal} &
\textbf{Refute} &
\textbf{Macro F1} \\
\midrule

Roberta \textsubscript{QLoRA} + ResNet50
  &  0.73 & 0.81 & 0.79 & 0.75 & 0.99 & 0.82 \\

Roberta \textsubscript{QLoRA} + ViT
  & 0.74 & 0.82 & 0.79 & 0.74 & 0.99 & 0.81 \\

DeBERTa \textsubscript{QLoRA} + ViT
  & 0.74 & 0.82 & 0.78 & 0.75 & 0.99 & 0.81 \\

DeBERTa \textsubscript{QLoRA} + ResNet50
  & 0.73 & 0.80 & 0.78 & 0.75 & 0.99 & 0.82 \\

SBERT \textsubscript{QLoRA} + ResNet50
  & 0.67 & 0.79 & 0.73 & 0.71 & 0.99 & 0.77 \\

LLaMA-3.1-8B \textsubscript{QLoRA} + ResNet50
  & \textbf{0.78} & \textbf{0.83} & \textbf{0.82} & \textbf{0.77} & \textbf{1.00} & \textbf{0.84} \\

LLaMA-3.1-8B \textsubscript{QLoRA} + ViT
  & 0.69 & 0.80 & 0.77 & 0.72 & 0.97 & 0.79 \\

Mistral7B \textsubscript{QLoRA} + ResNet50
  & \textbf{0.78} & \textbf{0.83} & \textbf{0.82} & \textbf{0.77} & \textbf{1.00} & \textbf{0.84} \\

Mistral7B \textsubscript{QLoRA} + ViT
  & 0.67 & 0.80 & 0.76 & 0.71 & 0.98 & 0.78 \\

\bottomrule[0.12em]
\end{tabular}
}
\caption{Class‐wise F1 scores and Macro F1 for model combinations \textbf{with contrastive head} on Factify 2 using QLoRA.}
\label{tab:factify2_results_qlora}
\end{table*}

%% file: tables/Mocheg_results_Cont.tex
\begin{table}[htbp]
\centering
\setlength{\tabcolsep}{3pt}
\renewcommand{\arraystretch}{1.15}

\begin{tabularx}{\linewidth}{
    >{\footnotesize\raggedright\arraybackslash}X   
    |c|c|c|
    >{\columncolor{gray!15}}c
}
\toprule[0.12em]
\textbf{Model With Contrastive Head} &
\textbf{Support} &
\textbf{NEI} &
\textbf{Refute} &
\shortstack{\textbf{Macro F1}} \\
\midrule

Roberta + ResNet50
 & 0.43 {\scriptsize($\pm$0.02)}
 & \textbf{0.37} {\scriptsize($\pm$0.02)}
 & 0.63 {\scriptsize($\pm$0.00)}
 & \textbf{0.48} {\scriptsize($\pm$0.01)} \\

Roberta + ViT
 & 0.46 {\scriptsize($\pm$0.04)}
 & 0.21 {\scriptsize($\pm$0.03)}
 & 0.63 {\scriptsize($\pm$0.01)}
 & \textbf{0.43} {\scriptsize($\pm$0.02)} \\

DeBERTa + ViT
 & 0.33 {\scriptsize($\pm$0.04)}
 & 0.17 {\scriptsize($\pm$0.03)}
 & \textbf{0.66} {\scriptsize($\pm$0.00)}
 & 0.39 {\scriptsize($\pm$0.01)} \\

DeBERTa + ResNet50
 & \textbf{0.51} {\scriptsize($\pm$0.02)}
 & 0.30 {\scriptsize($\pm$0.03)}
 & 0.63 {\scriptsize($\pm$0.01)}
 & \textbf{0.48} {\scriptsize($\pm$0.02)} \\

SBERT + ResNet50
 & 0.26 {\scriptsize($\pm$0.01)}
 & 0.05 {\scriptsize($\pm$0.03)}
 & 0.64 {\scriptsize($\pm$0.01)}
 & 0.37 {\scriptsize($\pm$0.01)} \\

Baseline 2
 & 0.41 
 & 0.17 
 & 0.51 
 & 0.36 \\

\bottomrule[0.12em]
\end{tabularx}
\caption{Mocheg results (3-way: Support, NEI, Refute). Each value is mean F1 $\pm$ std across seeds.}
\label{tab:mocheg_results}
\end{table}










%% file: tables/mocheg_QLoRA.tex
\begin{table}[htbp]
\centering
\setlength{\tabcolsep}{3pt}
\renewcommand{\arraystretch}{1.15}

\begin{tabularx}{\linewidth}{
    >{\footnotesize\raggedright\arraybackslash}X   
    |c|c|c|
    >{\columncolor{gray!15}}c
}
\toprule[0.12em]
\textbf{QLoRA Model With Contrastive Head} &
\textbf{Support} &
\textbf{NEI} &
\textbf{Refute} &
\shortstack{\textbf{Macro F1}} \\
\midrule
Roberta \textsubscript{QLoRA} + ResNet50
  & 0.40
  & 0.31
  & 0.54
  & 0.42 \\

Roberta \textsubscript{QLoRA} + ViT
  & 0.37
  & 0.32
  & 0.51
  & 0.40 \\

DeBERTa \textsubscript{QLoRA} + ViT
  & 0.46
  & 0.38
  & 0.57
  & 0.47\\

DeBERTa \textsubscript{QLoRA} + ResNet50
  & 0.46
  & 0.39
  & 0.57
  & 0.47 \\

SBERT \textsubscript{QLoRA} + ResNet50
  & 0.31
  & 0.29
  & 0.52
  & 0.37 \\

LLaMA-3.1-8B \textsubscript{QLoRA} + ResNet50
  & 0.43
  & 0.43
  & 0.58
  & 0.48 \\
LLaMA-3.1-8B \textsubscript{QLoRA} + ViT
  & 0.15
  & 0.33
  & 0.28
  & 0.25 \\
Mistral7B \textsubscript{QLoRA} + ResNet50
  & 0.46
  & \textbf{0.44}
  & 0.58
  & \textbf{0.49} \\
Mistral7B \textsubscript{QLoRA} + ViT
  & 0.13
  & 0.32
  & 0.30
  & 0.25 \\

\bottomrule[0.12em]
\end{tabularx}
\caption{Mocheg results (3-way: Support, NEI, Refute). Each value is mean F1 $\pm$ std across seeds.}
\label{tab:mocheg_results_qlora}
\end{table}

%% file: tables/mocheg_result_no_contrastive_head.tex
\begin{table}[htbp]
\centering
\setlength{\tabcolsep}{3pt}
\renewcommand{\arraystretch}{1.15}

\begin{tabularx}{\linewidth}{
    >{\footnotesize\raggedright\arraybackslash}X   
    |c|c|c|
    >{\columncolor{gray!15}}c
}
\toprule[0.12em]
\textbf{Model With Contrastive Head} &
\textbf{Support} &
\textbf{NEI} &
\textbf{Refute} &
\shortstack{\textbf{Macro F1} \\ \textbf{Score}} \\
\midrule

Roberta + ResNet50
 & 0.39 {\scriptsize($\pm$0.02)}
 & 0.34 {\scriptsize($\pm$0.02)}
 & 0.57 {\scriptsize($\pm$0.01)}
 & 0.43 {\scriptsize($\pm$0.00)} \\
Roberta + ViT
 & 0.36 {\scriptsize($\pm$0.03)}
 & 0.18 {\scriptsize($\pm$0.02)}
 & 0.55 {\scriptsize($\pm$0.01)}
 & 0.36 {\scriptsize($\pm$0.01)} \\

DeBERTa + ViT
 & 0.32 {\scriptsize($\pm$0.02)}
 & 0.12 {\scriptsize($\pm$0.03)}
 & 0.58 {\scriptsize($\pm$0.00)}
 & 0.34 {\scriptsize($\pm$0.01)} \\

DeBERTa + ResNet50
 & 0.43 {\scriptsize($\pm$0.03)}
 & 0.27 {\scriptsize($\pm$0.02)}
 & 0.59 {\scriptsize($\pm$0.01)}
 & 0.40 {\scriptsize($\pm$0.02)} \\
SBERT + ResNet50
 & 0.21 {\scriptsize($\pm$0.01)}
 & 0.06 {\scriptsize($\pm$0.02)}
 & 0.58 {\scriptsize($\pm$0.01)}
 & 0.28 {\scriptsize($\pm$0.01)}  \\
 
Baseline 2
 & 0.41 
 & 0.17 
 & 0.51 
 & 0.36 \\

\bottomrule[0.12em]
\end{tabularx}
\caption{Mocheg results (3-way: Support, NEI, Refute) without contrasive head. Each value is mean F1 $\pm$ std across seeds.}
\label{tab:mocheg_results_without_contrastive_head}
\end{table}

%% file: tables/factify_three_way.tex
\begin{table}[htbp]
\centering
\setlength{\tabcolsep}{3pt}
\renewcommand{\arraystretch}{1.15}


\begin{tabularx}{\linewidth}{
    >{\footnotesize\raggedright\arraybackslash}X   
    |c|c|c|
    >{\columncolor{gray!15}}c
}
\toprule[0.12em]
\textbf{Model With Contrastive Head} &
\textbf{Support} &
\textbf{NEI} &
\textbf{Refute} &
\shortstack{\textbf{Macro F1}} \\
\midrule

Roberta + ResNet50
 & 0.84 {\scriptsize($\pm$0.01)} & 0.83 {\scriptsize($\pm$0.01)} & 0.99 {\scriptsize($\pm$0.00)} & 0.89 {\scriptsize($\pm$0.01)} \\
Roberta + ViT
 & 0.85 {\scriptsize($\pm$0.01)} & 0.84 {\scriptsize($\pm$0.01)} & 1.00 {\scriptsize($\pm$0.00)} & 0.90 {\scriptsize($\pm$0.00)} \\

DeBERTa + ViT
 & 0.86 {\scriptsize($\pm$0.01)} & 0.85 {\scriptsize($\pm$0.01)} & 1.00 {\scriptsize($\pm$0.00)} & 0.90 {\scriptsize($\pm$0.01)}  \\

DeBERTa + ResNet50
 & 0.85 {\scriptsize($\pm$0.00)} & 0.84 {\scriptsize($\pm$0.00)} & 1.00 {\scriptsize($\pm$0.00)} & 0.90 {\scriptsize($\pm$0.00)} \\
SBERT + ResNet50
 &0.83 {\scriptsize($\pm$0.00)} & 0.82 {\scriptsize($\pm$0.00)} & 1.00 {\scriptsize($\pm$0.00)} & 0.88 {\scriptsize($\pm$0.00)}   \\
 
baseline 2
 &0.75  &  0.76  &  0.99  & 0.84  \\

\bottomrule[0.12em]
\end{tabularx}
\caption{Factify 2 (3-way: Support, NEI, Refute) contrastive head. Each value is mean F1 $\pm$ std across seeds.}
\label{tab:factify_three_way_label}
\end{table}

%% file: tables/factify_three_way_noncont.tex
\begin{table}[htbp]
\centering
\setlength{\tabcolsep}{3pt}
\renewcommand{\arraystretch}{1.15}

\begin{tabularx}{\linewidth}{
    >{\footnotesize\raggedright\arraybackslash}X   
    |c|c|c|
    >{\columncolor{gray!15}}c
}
\toprule[0.12em]
\textbf{Model Without Contrastive Head} &
\textbf{Support} &
\textbf{NEI} &
\textbf{Refute} &
\shortstack{\textbf{Macro F1}} \\
\midrule

Roberta + ResNet50
 & 0.81 {\scriptsize($\pm$0.00)} & 0.79 {\scriptsize($\pm$0.01)} & 1.00{\scriptsize($\pm$0.00)} & 0.86 {\scriptsize($\pm$0.02)} \\
Roberta + ViT
 & 0.81 {\scriptsize($\pm$0.00)} & 0.81 {\scriptsize($\pm$0.01)} & 1.00 {\scriptsize($\pm$0.00)} & 0.87{\scriptsize($\pm$0.00)} \\

DeBERTa + ViT
 & 0.82 {\scriptsize($\pm$0.00)} & 0.80 {\scriptsize($\pm$0.01)} & 1.00 {\scriptsize($\pm$0.00)} & 0.87 {\scriptsize($\pm$0.00)}  \\

DeBERTa + ResNet50
 & 0.82 {\scriptsize($\pm$0.02)} & 0.81 {\scriptsize($\pm$0.03)} & 1.00 {\scriptsize($\pm$0.00)} & 0.88 {\scriptsize($\pm$0.01)} \\
SBERT + ResNet50
 &0.80 {\scriptsize($\pm$0.00)} & 0.79 {\scriptsize($\pm$0.01)} & 1.00 {\scriptsize($\pm$0.00)} & 0.86 {\scriptsize($\pm$0.00)}   \\

baseline 2
 &0.75  &  0.76  &  0.99  & 0.84  \\

\bottomrule[0.12em]
\end{tabularx}
\caption{Factify 2 (3-way: Support, NEI, Refute) without contrastive head. Each value is the mean F1 $\pm$ std across seeds.}
\label{tab:factify_three_way_label_non_cont}
\end{table}

%% file: 70ErrorAnalysis.tex
\section{ Ablation Studies:}


\begin{itemize}
    \item \textbf{Modality Masking:} To assess reliance on visual signals, we masked either the claim or document image while keeping text and OCR intact. Performance drops were modest: the full model (RoBERTa + ResNet50, contrastive head) fell from 0.84 → 0.75 when masking the claim image (–0.09) and 0.84 → 0.73 when masking the document image (–0.11). This shows that MultiCheck degrades gracefully and that text + OCR provide the primary discriminative signal, while images act as complementary evidence, particularly for document-level inconsistencies.

    \item \textbf{OCR Noise Robustness:} We injected $\approx$ 20\% synthetic OCR noise into both claim and document text to evaluate noise robustness. Performance remained effectively unchanged (0.84 → 0.84,  $\delta$ $\approx$ 0.0037), indicating that MultiCheck is highly stable under OCR corruption. The contrastive objective helps stabilize the model, preventing degradation under noisy text.

    \item \textbf{Hyper-parameter sensitivity of contrastive loss} We performed a grid sweep over $\lambda$ and $\tau$ reported in Table \ref{tab:lambdatau}.  Performance is stable in the 0.82–0.83 macro-F1 band,  with a shallow optimum around $(\lambda,\tau)\!\approx\!(0.1,0.1)$,  confirming robustness of the dual-loss design.

\end{itemize}

\input{tables/sweep}

\section{Error analysis:}

We analysed the prediction mismatches between our approach and the baseline to better understand model behaviour. As shown in Figure~\ref{fig:qualitative_sample}, the baseline relied only on text, predicting \textit{Insufficient Text}. However, our model used the OCR credit “Helen Sloan/HBO” to recognize the image as a licensed promotional photograph,  details suggesting the image does not contribute new factual content, and correctly predicted \textit{Insufficient Multimodal}. By recognizing the lack of substantive support in text and image, our model rightly predicted \textit{Insufficient Multimodal}. This highlights the role of OCR-extracted image sources in improving factual reasoning. Additional examples are discussed below:
\vspace{2mm}

\noindent\textbf{Additional examples of qualitative analysis:}
\label{quality_error}

\noindent Here, we have presented the dataset-referenced examples comparing our model with the baseline, showing how OCR cues such as credits, agency marks, or overlays can clarify or complicate the multimodal verification.


\noindent\textbf{Example A: When MultiCheck outperforms the baseline}
\begin{itemize}
    \item \textbf{ID 4681:} The baseline, relying only on text, misclassified “Made in India vaccines” as \textit{Support\_Text}. In contrast, our model's multimodal fusion leveraged the OCR cue “COLO STORAE” (cold storage) to detect ambiguity and correctly predicted \textit{Insufficient\_Text}. In Figure \ref{fig:qualitative_sample_4681}, this illustrates how even noisy OCR can reveal context missed by text-only models.
    
    %
    \item \textbf{ID 6968:} The baseline predicted \textit{Support\_Text} based on name and context matches in text (e.g., “Governor Jagdeep Dhankar”). In contrast, our model used the OCR cue “ANI” to identify the image as a generic press photo and, combining it with weak textual alignment, correctly predicted \textit{Insufficient\_Multimodal}. This demonstrates its ability to dismiss superficial visual signals.

\end{itemize}

\noindent\textbf{Example B: When baseline outperforms MultiCheck}
\begin{itemize}
    \item \textbf{ID 7171:} In this illustrated in Figure \ref{fig:qualitative_sample_7171}, our model was overly cautious. The OCR tag “ANI” led it to dismiss the image as generic, resulting in an \textit{Insufficient\_Multimodal} prediction. However, the textual portion “will lay foundation stone” is a direct and verifiable event announcement, and the image provided valid context. The baseline correctly predicted \textit{Support\_Multimodal}, highlighting that OCR cues can sometimes mislead when visual content, though generic, remains contextually relevant.

\end{itemize}

\subsection{Quantized variants and memory footprint.}
\label{Quantized_version_discussion}

To study the deployability of MultiCheck under strict GPU budgets, we evaluated both the initially configured variants and their 4-bit quantized counterparts. For the RoBERTa / DeBERTa / SBERT configurations, we applied NF4 QLoRA to the text encoder while keeping the ResNet50 or ViT image encoders unchanged. Tables 16 and 18 report the peak GPU memory usage during training at a batch size of 32, as measured by \texttt{nvidia-smi}. On Factify-2, quantization reduces memory usage by roughly \textbf{30–60}\% across all backbones. The macro-F1 drop is negligible or within a few points, indicating that MultiCheck’s fusion and contrastive heads remain effective even with low-precision text encoders. We further replaced the text encoder with larger causal language models—LLaMA-3.1-8B and Mistral-7B—and trained them using 4-bit NF4 QLoRA. As shown in Tables 17 and 19, the quantized LLM-based variants fit within 13–15,GB of GPU memory while incurring only small performance degradation on Factify-2 (negligible for LLaMA/Mistral+ResNet50 and at most a 5–6 macro-F1 point drop for ViT-based models). On Mocheg, the quantized LLaMA/Mistral+ResNet50 models also maintain competitive performance with negligible loss, whereas the ViT-based variants show a larger drop, suggesting that a higher-capacity text backbone is more beneficial when paired with a stable visual encoder. Overall, these results demonstrate that MultiCheck can leverage modern LLM backbones in an energy- and memory-efficient manner, making it feasible to deploy multi-billion-parameter text encoders for multimodal fact-checking on a single commodity GPU.

%% file: tables/sweep.tex
\begin{table}[h]
\centering
\begin{tabular}{cccc}
\toprule
$\lambda$ & $\tau$ & \textbf{Test M-F1} & \textbf{Accuracy (\%)}\\
\midrule
0.10 & 0.10 & 0.84 & 84.08\\
0.10 & 0.05 & 0.83 & 82.65\\
0.05 & 0.10 & 0.83 & 82.52\\
0.05 & 0.20 & 0.83 & 81.77\\
0.05 & 0.05 & 0.83 & 82.50\\
\bottomrule
\end{tabular}
\caption{Macro F1 and accuracy across $\lambda,\tau$ combinations.}
\label{tab:lambdatau}
\end{table}

%% file: 80Conclusion.tex
\section{Conclusion:}
This paper introduced MultiCheck, a unified framework for fine-grained multimodal fact-checking that jointly reasons over text, images, and OCR signals. By combining structured representations from pre-trained encoders with a relational fusion module (element-wise difference and product) and a supervised contrastive objective, MultiCheck captures subtle agreements and contradictions between claims and evidence. Across Factify-2 and Mocheg, our models consistently outperform strong baselines, with especially large gains on challenging “Insufficient” and “Refute” cases. 

%% file: 90appendix.tex
\section*{Appendix}

\input{tables/significance_contrastive}
\input{tables/significance}
\input{tables/factify_result_no_head}

\begin{figure}[ht]
\centering
\includegraphics[width=\columnwidth]{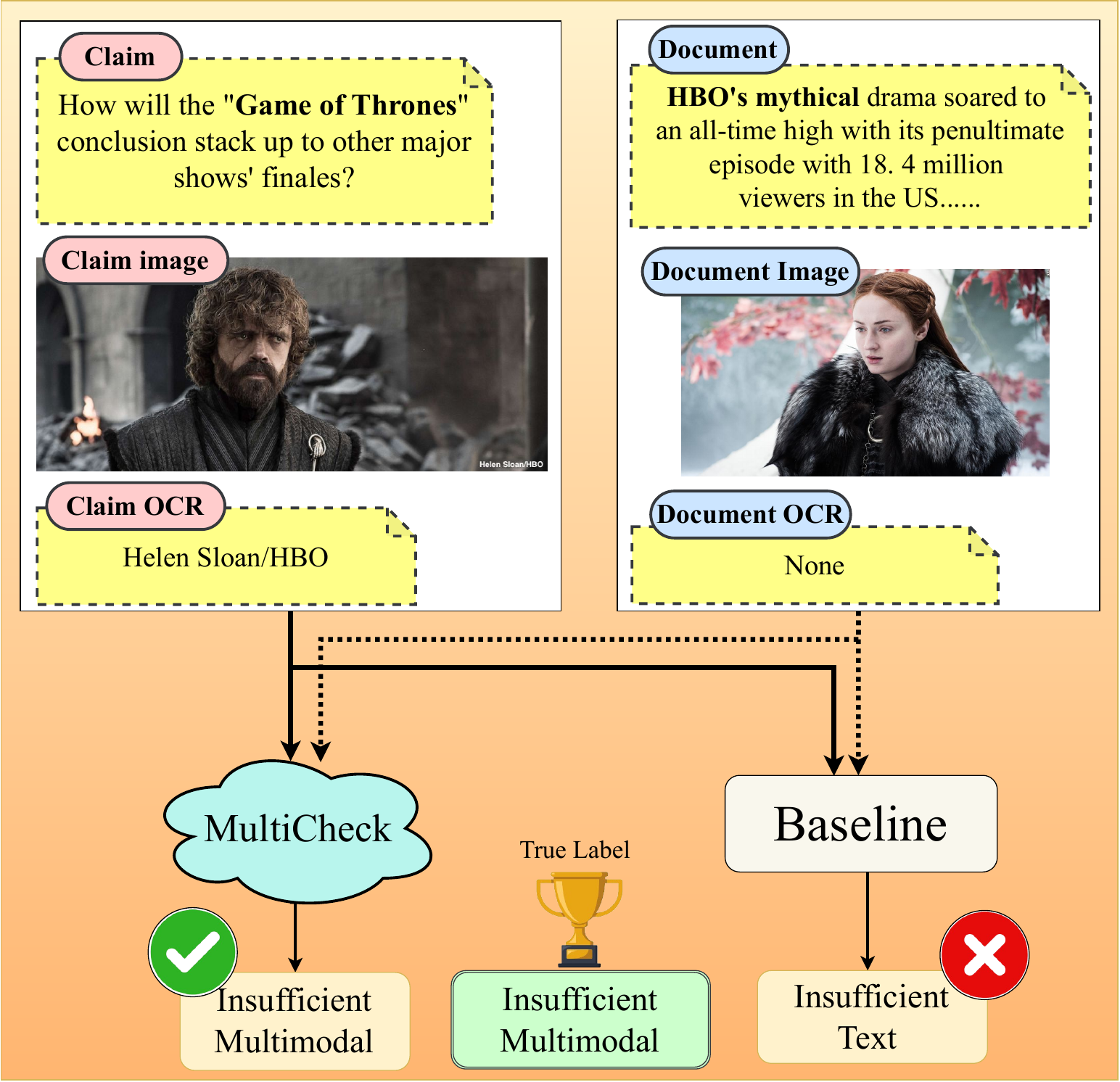}
\caption{Example of qualitative analysis, Sample from the dataset.}
\label{fig:qualitative_sample}
\vspace{-2mm}
\end{figure}

\begin{figure}[ht]
\centering
\includegraphics[width=\columnwidth]{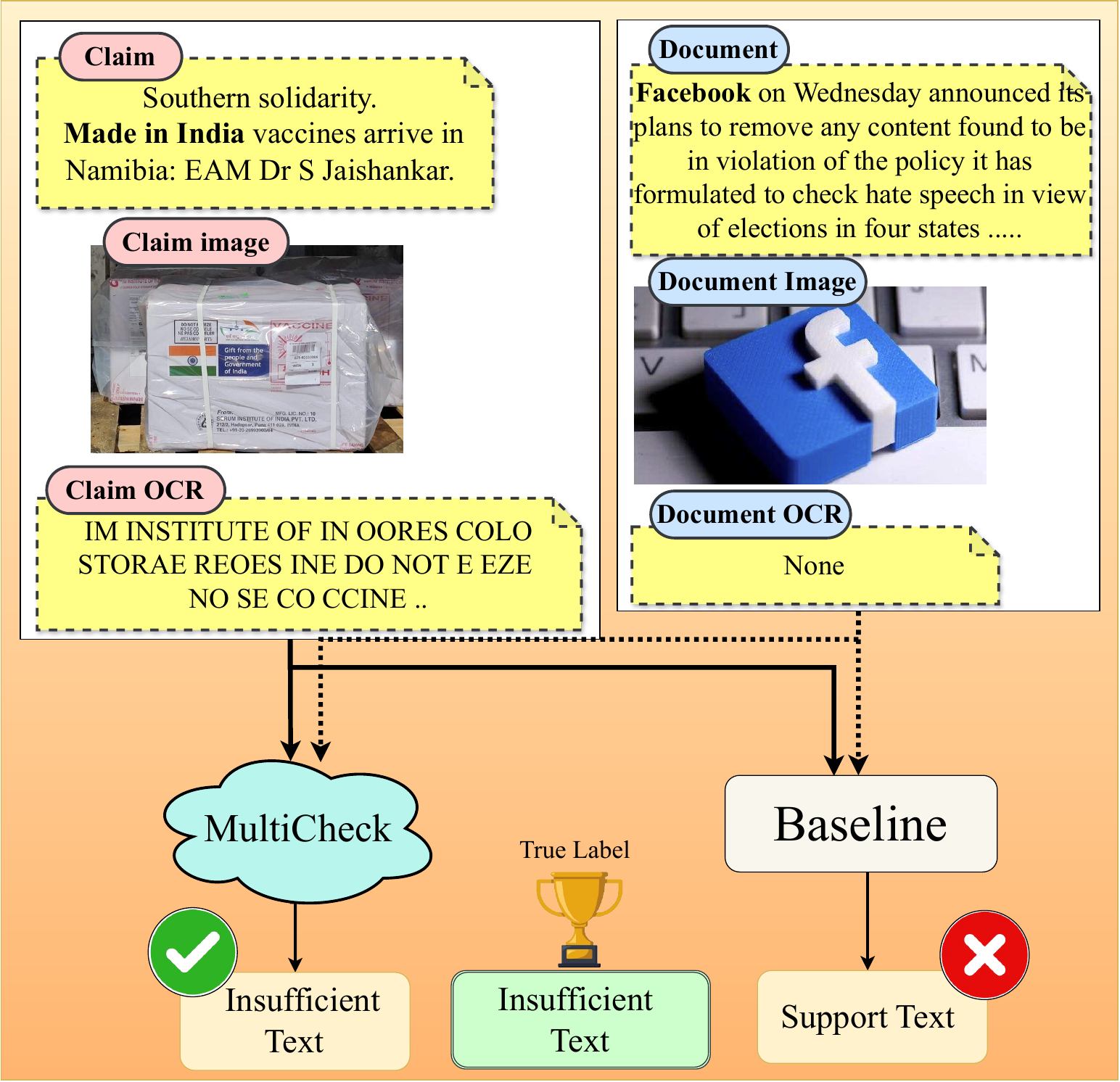}
\caption{Example of qualitative analysis of ID-4681, Sample from the dataset.}
\label{fig:qualitative_sample_4681}
\vspace{-2mm}
\end{figure}

\begin{figure}[ht]
\centering
\includegraphics[width=\columnwidth]{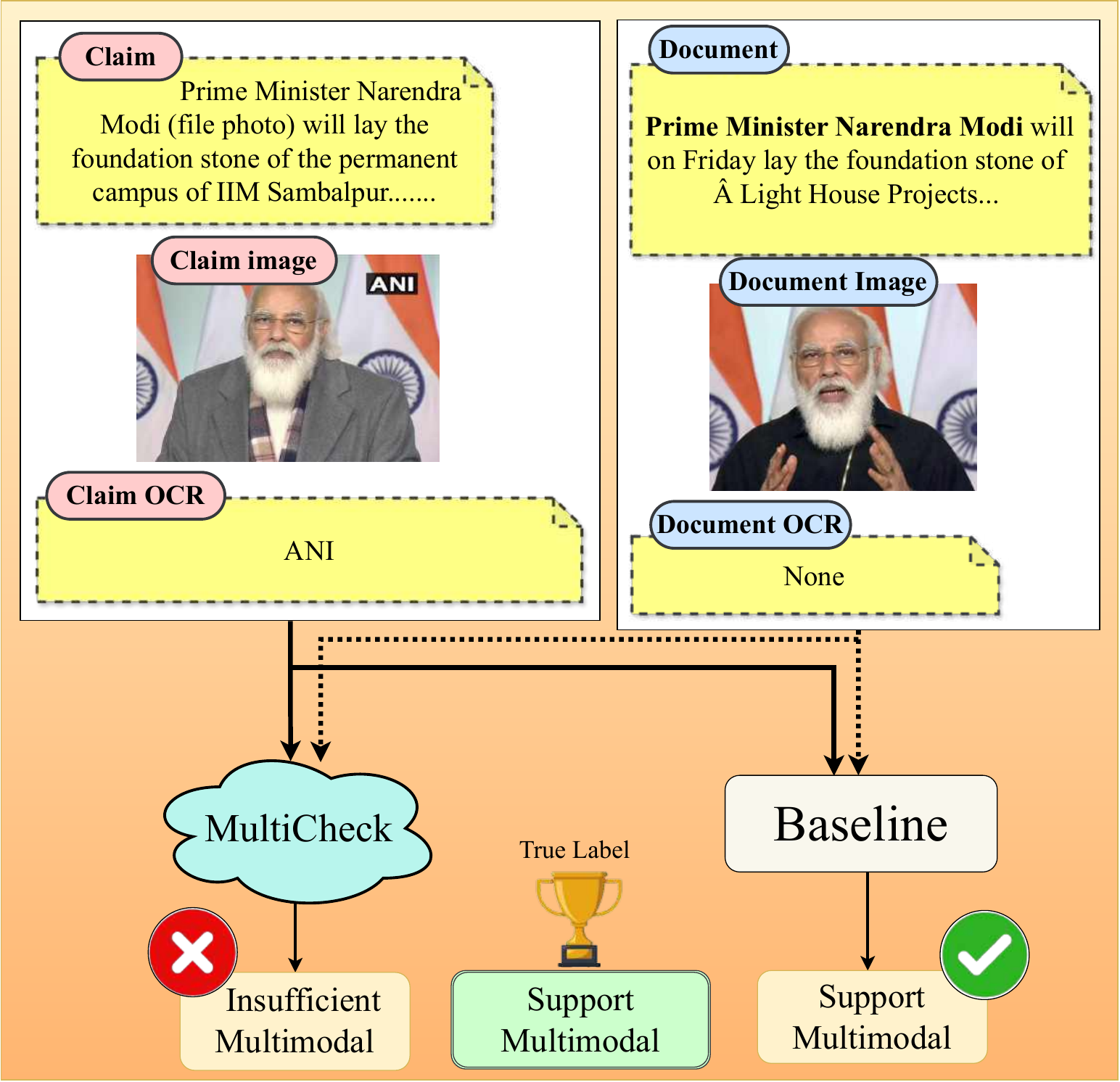}
\caption{Example of qualitative analysis of ID-7171, Sample from the dataset.}
\label{fig:qualitative_sample_7171}
\vspace{-2mm}
\end{figure}

\input{tables/footprint_factify}
\input{tables/footprint_mocheg}

\begin{table}[htbp]
\centering
\setlength{\tabcolsep}{4pt}
\renewcommand{\arraystretch}{1.15}

\begin{tabularx}{\columnwidth}{
    >{\footnotesize\raggedright\arraybackslash}X
    |c
    |c
    |>{\columncolor{gray!15}}c
}
\toprule[0.12em]
\textbf{Models With Contrastive Head Memory footprint} &
\shortstack{\textbf{Original} \\ \textbf{Models}}  &
\shortstack{\textbf{Quantized} \\ \textbf{Models}} &
\shortstack{\textbf{Reduction} \\ \textbf{in Memory}} \\
\midrule

Roberta + ResNet50
  & $\approx$24 GB
  & $\approx$12 GB
  & \textbf{50}\% \\

Roberta + ViT
  & $\approx$20 GB
  & $\approx$12 GB
  & \textbf{40}\% \\

DeBERTa + ViT
  & $\approx$28 GB 
  & $\approx$20 GB 
  & \textbf{29}\% \\

DeBERTa + ResNet50
  & $\approx$27 GB
  & $\approx$18 GB
  & \textbf{33}\% \\

SBERT + ResNet50
  & $\approx$12 GB
  & $\approx$5 GB
  & \textbf{58}\% \\

\bottomrule[0.12em]
\end{tabularx}
\caption{Model Memory footprint comparison between quantized and original versions on GPU computation for factify}
\label{tab:footprint_comparison}
\end{table}


\begin{table}[htbp]
\centering
\setlength{\tabcolsep}{4pt}
\renewcommand{\arraystretch}{1.15}

\begin{tabularx}{\columnwidth}{
    >{}X
    |>{\columncolor{gray!15}}c
    |c
}
\toprule[0.12em]
\textbf{Models Memory footprint } &
\shortstack{\textbf{Quantized} \\ \textbf{Version}} &
\shortstack{\textbf{Performance} \\ \textbf{Degradation}} \\
\midrule

LLaMA-3.1-8B + ResNet50
  & $\approx$\textbf{14} GB
  &  Negligible\\

LLaMA-3.1-8B + ViT
  & $\approx$14 GB
  & by 5 points\\

Mistral7B + ViT
  & $\approx$15 GB
  &  by 6 points\\

Mistral7B + ResNet50
  & $\approx$\textbf{15} GB
  &  Negligible\\

\bottomrule[0.12em]
\end{tabularx}
\caption{Model Memory footprint of quantized LLM-based models on GPU computation for factify}
\label{tab:footprint_comparison_LLMs}
\end{table}

%% file: tables/significance_contrastive.tex
\newcommand{\cmark}{\checkmark}

\begin{table*}[htbp]
\centering
\setlength{\tabcolsep}{6pt}
\renewcommand{\arraystretch}{1.2}

\resizebox{\textwidth}{!}{
\begin{tabular}{l|c|c|>{\columncolor{green!7.5}}c|c|c|c}
\toprule[0.12em]
\textbf{Models Without Contrastive Head} &
\makecell{\textbf{McNemar’s}\\\textbf{$\chi^2$}} &
\makecell{\textbf{McNemar’s}\\\textbf{p-value}} &
\makecell{\textbf{Significance}\\\textbf{at $\alpha=0.05$}} &
\makecell{\textbf{Bowker’s}\\\textbf{$\chi^2$ (df=10)}} &
\makecell{\textbf{Bowker’s}\\\textbf{p-value}} &
\makecell{\textbf{Reject}\\\textbf{symmetry}} \\
\midrule

Roberta + ResNet50 & 1129.91 & $\ll0.05$ & \cmark & 75.64 & $3.56\times10^{-12}$ & \cmark \\
Roberta + ViT      & 1243.45 & $\ll0.05$ & \cmark & 140.44 & $0.00$              & \cmark \\
DeBERTa + ViT      & 965.16  & $\ll0.05$ & \cmark & 122.93 & $0.00$              & \cmark \\
DeBERTa + ResNet50 & 1195.33 & $\ll0.05$ & \cmark & 153.76 & $0.00$              & \cmark \\
SBERT + ResNet50   & 891.88  & $\ll0.05$ & \cmark & 100.06 & $0.00$              & \cmark \\

\midrule
\multicolumn{7}{l}{\textbf{Models With Contrastive Head}} \\
\midrule

Roberta + ResNet50 & 1238.56 & $\ll0.05$ & \cmark & 91.85  & $2.33\times10^{-15}$ & \cmark \\
Roberta + ViT      & 1333.14 & $\ll0.05$ & \cmark & 132.10 & $0.00$               & \cmark \\
DeBERTa + ViT      & 1480.57 & $\ll0.05$ & \cmark & 118.33 & $0.00$               & \cmark \\
DeBERTa + ResNet50 & 1217.24 & $\ll0.05$ & \cmark & 166.16 & $0.00$               & \cmark \\
SBERT + ResNet50   & 1070.45 & $\ll0.05$ & \cmark & 82.03  & $2.00\times10^{-13}$ & \cmark \\

\bottomrule[0.12em]
\end{tabular}
}

\caption{Significance‐test results comparing the multimodal model against the baseline on Factify 2.}
\label{tab:significance_contrastive_tests}
\end{table*}

%% file: tables/factify_result_no_head.tex
\begin{table*}[htbp]
\centering
\setlength{\tabcolsep}{6pt}
\renewcommand{\arraystretch}{1.2}

\resizebox{\linewidth}{!}{
\begin{tabular}{l|c|c|c|c|c|>{\columncolor{gray!15}}c}
\toprule[0.12em]
\textbf{Models Without Contrastive Head} &
\textbf{Support Text} &
\textbf{Support Multimodal} &
\textbf{Insufficient Text} &
\textbf{Insufficient Multimodal} &
\textbf{Refute} &
\textbf{Macro F1} \\
\midrule

Roberta + ResNet50
  & 0.74\unc{0.01}
  & \bfseries0.82\unc{0.01}
  & 0.78\unc{0.01}
  & \bfseries0.75\unc{0.01}
  & 0.99\unc{0.01}
  & \bfseries0.82\unc{0.01} \\

Roberta + ViT
  & \bfseries0.75\unc{0.01}
  & \bfseries0.82\unc{0.01}
  & \bfseries0.79\unc{0.01}
  & \bfseries0.75\unc{0.01}
  & \bfseries1.00\unc{0.00}
  & \bfseries0.82\unc{0.00} \\

DeBERTa + ViT
  & 0.74\unc{0.01}
  & \bfseries0.82\unc{0.00}
  & 0.78\unc{0.01}
  & \bfseries0.75\unc{0.01}
  & \bfseries1.00\unc{0.00}
  & \bfseries0.82\unc{0.00} \\

DeBERTa + ResNet50
  & 0.74\unc{0.01}
  & 0.81\unc{0.01}
  & 0.78\unc{0.01}
  & 0.74\unc{0.01}
  & \bfseries1.00\unc{0.00}
  & 0.81\unc{0.00} \\

SBERT + ResNet50
  & 0.71\unc{0.03}
  & 0.81\unc{0.02}
  & 0.75\unc{0.02}
  & 0.74\unc{0.02}
  & 0.99\unc{0.00}
  & 0.78\unc{0.05} \\

\bottomrule[0.12em]
\end{tabular}
}
\caption{Class‐wise F1 scores and Macro F1 for various model combinations on the Factify 2 dataset (no contrastive head). Each value is mean F1 $\pm$std across seeds.}
\label{tab:factify2_results_no_contrastive}
\end{table*}

%% file: tables/footprint_factify.tex
\begin{table}[htbp]
\centering
\setlength{\tabcolsep}{4pt}
\renewcommand{\arraystretch}{1.15}

\begin{tabularx}{\columnwidth}{
    >{\footnotesize\raggedright\arraybackslash}X
    |c
    |c
    |>{\columncolor{gray!15}}c
}
\toprule[0.12em]
\textbf{Models With Contrastive Head Memory footprint} &
\shortstack{\textbf{Original} \\ \textbf{Models}}  &
\shortstack{\textbf{Quantized} \\ \textbf{Models}} &
\shortstack{\textbf{Reduction} \\ \textbf{in Memory}} \\
\midrule

Roberta + ResNet50
  & $\approx$24 GB
  & $\approx$12 GB
  & \textbf{50}\% \\

Roberta + ViT
  & $\approx$20 GB
  & $\approx$12 GB
  & \textbf{40}\% \\

DeBERTa + ViT
  & $\approx$28 GB 
  & $\approx$20 GB 
  & \textbf{29}\% \\

DeBERTa + ResNet50
  & $\approx$27 GB
  & $\approx$18 GB
  & \textbf{33}\% \\

SBERT + ResNet50
  & $\approx$12 GB
  & $\approx$5 GB
  & \textbf{58}\% \\

\bottomrule[0.12em]
\end{tabularx}
\caption{Model Memory footprint comparison between quantized and original versions on GPU computation for factify}
\label{tab:footprint_comparison}
\end{table}


\begin{table}[htbp]
\centering
\setlength{\tabcolsep}{4pt}
\renewcommand{\arraystretch}{1.15}

\begin{tabularx}{\columnwidth}{
    >{}X
    |>{\columncolor{gray!15}}c
    |c
}
\toprule[0.12em]
\textbf{Models Memory footprint } &
\shortstack{\textbf{Quantized} \\ \textbf{Version}} &
\shortstack{\textbf{Performance} \\ \textbf{Degradation}} \\
\midrule

LLaMA-3.1-8B + ResNet50
  & $\approx$\textbf{14} GB
  &  Negligible\\

LLaMA-3.1-8B + ViT
  & $\approx$14 GB
  & by 5 points\\

Mistral7B + ViT
  & $\approx$15 GB
  &  by 6 points\\

Mistral7B + ResNet50
  & $\approx$\textbf{15} GB
  &  Negligible\\

\bottomrule[0.12em]
\end{tabularx}
\caption{Model Memory footprint of quantized LLM-based models on GPU computation for factify}
\label{tab:footprint_comparison_LLMs}
\end{table}

%% file: tables/footprint_mocheg.tex
\begin{table}[htbp]
\centering
\setlength{\tabcolsep}{4pt}
\renewcommand{\arraystretch}{1.15}

\begin{tabularx}{\columnwidth}{
    >{\footnotesize\raggedright\arraybackslash}X
    |c
    |c
    |>{\columncolor{gray!15}}c
}
\toprule[0.12em]
\textbf{Models With Contrastive Head Memory footprint} &
\shortstack{\textbf{Original} \\ \textbf{Models}}  &
\shortstack{\textbf{Quantized} \\ \textbf{Models}} &
\shortstack{\textbf{Reduction} \\ \textbf{in Memory}} \\
\midrule

Roberta + ResNet50
  & $\approx$12 GB
  & $\approx$5 GB
  & \textbf{58}\% \\

Roberta + ViT
  & $\approx$13 GB
  & $\approx$7 GB
  & \textbf{46}\% \\

DeBERTa + ViT
  & $\approx$13 GB 
  & $\approx$5 GB 
  & \textbf{62}\% \\

DeBERTa + ResNet50
  & $\approx$15 GB
  & $\approx$7 GB
  & \textbf{53}\% \\

SBERT + ResNet50
  & $\approx$12 GB
  & $\approx$4 GB
  & \textbf{67}\% \\

\bottomrule[0.12em]
\end{tabularx}
\caption{Model Memory footprint comparison between quantized and original versions on GPU computation for Mocheg}
\label{tab:footprint_comparison_Mocheg}
\end{table}


\begin{table}[htbp]
\centering
\setlength{\tabcolsep}{4pt}
\renewcommand{\arraystretch}{1.15}

\begin{tabularx}{\columnwidth}{
    >{}X
    |>{\columncolor{gray!15}}c
    |c
}
\toprule[0.12em]
\textbf{Models Memory footprint } &
\shortstack{\textbf{Quantized} \\ \textbf{Version}} &
\shortstack{\textbf{Performance} \\ \textbf{Degradation}} \\
\midrule

LLaMA-3.1-8B + ResNet50
  & $\approx$\textbf{13} GB
  &  Negligible\\

LLaMA-3.1-8B + ViT
  & $\approx$13 GB
  & by 23 points\\

Mistral7B + ViT
  & $\approx$10 GB
  &  by 23 points\\

Mistral7B + ResNet50
  & $\approx$\textbf{10} GB
  &  Negligible\\

\bottomrule[0.12em]
\end{tabularx}
\caption{Model Memory footprint of quantized LLM-based models on GPU computation for Mocheg}
\label{tab:footprint_comparison_LLMs_Mocheg}
\end{table}